%% file: main.tex
\documentclass[10pt,twocolumn,letterpaper]{article}

\usepackage[pagenumbers]{iccv} 

\input{preamble}

\definecolor{iccvblue}{rgb}{0.21,0.49,0.74}
\usepackage[pagebackref,breaklinks,colorlinks,allcolors=iccvblue]{hyperref}

\usepackage{lipsum}
\usepackage{multirow}
\usepackage{multicol}
\usepackage{colortbl}
\usepackage{xcolor}
\usepackage{subcaption}
\usepackage{amssymb}
\usepackage{pifont}
\usepackage{blindtext}
\usepackage{array}
\usepackage{cellspace}
\usepackage{booktabs}
\usepackage{graphicx}
\usepackage{relsize}
\usepackage{listings}
\usepackage{pifont} 
\usepackage{xcolor}   

\newcommand{\cmark}{\textcolor{ForestGreen}{\ding{51}}} 
\newcommand{\xmark}{\textcolor{red}{\ding{55}}}   

\def\paperID{2354} 
\def\confName{ICCV}
\def\confYear{2025}

\title{URECA \makebox[14pt][r]{\raisebox{-0.55ex}{\includegraphics[scale=0.1]{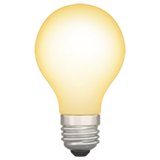}}}: Unique Region Caption Anything}

\author{
Sangbeom Lim$^{1*}$ \qquad Junwan Kim$^{2*}$ \qquad Heeji Yoon$^{3}$ \qquad
Jaewoo Jung$^{3}$ \qquad Seungryong Kim$^{3\dagger}$ \\[5pt]
$^{1}$Korea University \qquad $^{2}$Yonsei University \qquad $^{3}$KAIST AI
\\
[5pt]
{\tt \href{https:/cvlab-kaist.github.io/URECA}{https://cvlab-kaist.github.io/URECA}}
\vspace{-10pt}
}

\begin{document}
\twocolumn[{%
\renewcommand\twocolumn[1][]{#1}%
\maketitle
\centering
    \includegraphics[width=\linewidth]{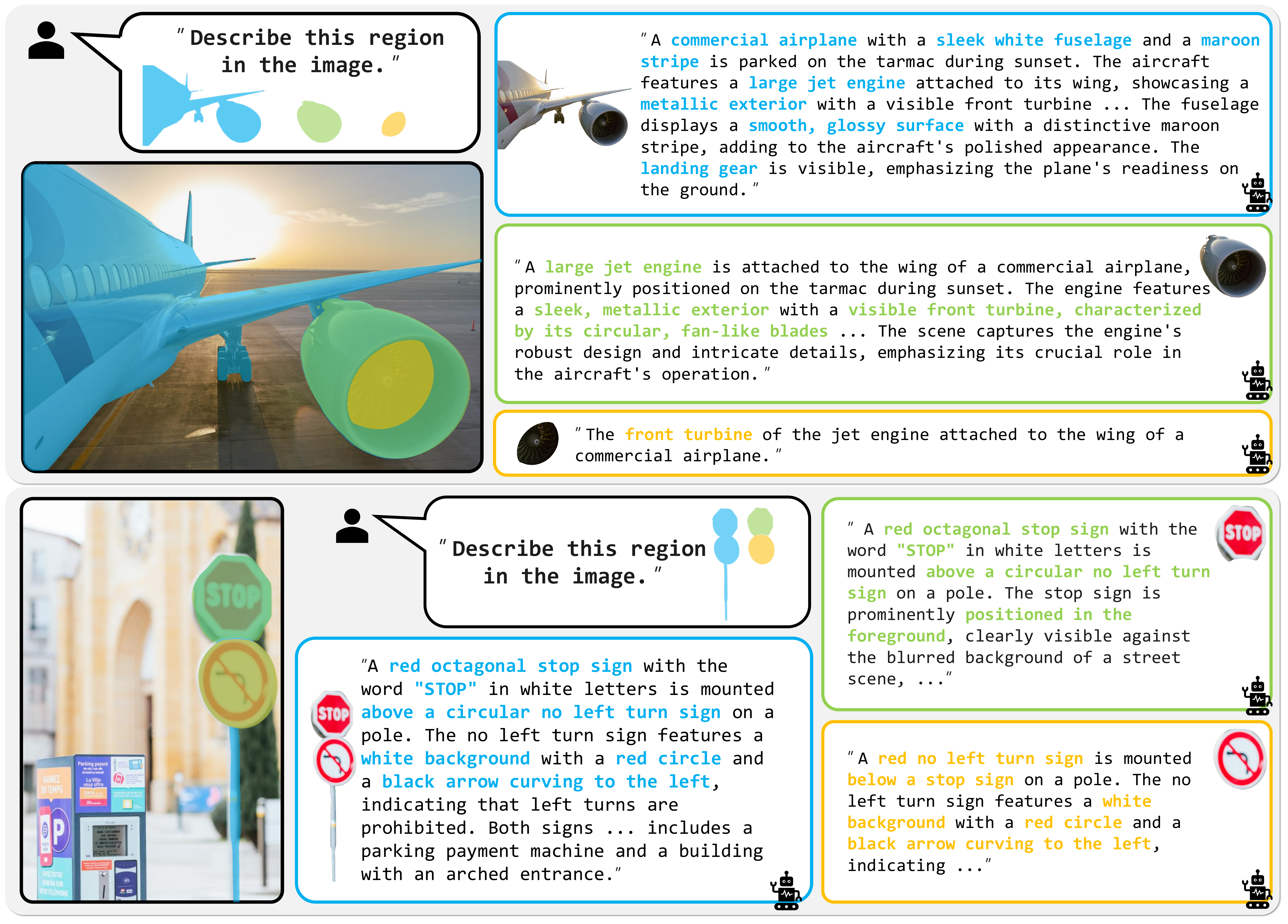}
\vspace{-10pt}
\captionof{figure}{\textbf{Unique Region Caption Anything.} We introduce \ours, a novel region-level captioning dataset designed to ensure caption uniqueness and support multi-granularity regions. Each caption in our benchmark is uniquely mapped to its corresponding region, capturing distinctive attributes that differentiate it from surrounding areas. Moreover, we show that our proposed model trained on our dataset effectively generates unique captions for regions at any level of granularity.
\vspace{1em}
}
\label{fig:teaser}
}]

\def\thefootnote{*}\footnotetext{These authors contributed equally.}\def\thefootnote{\arabic{footnote}}
\def\thefootnote{$\dagger$}\footnotetext{Corresponding author.}\def\thefootnote{\arabic{footnote}}

\input{sec/0_abstract}    
\input{sec/1_first_revised_intro}
\input{sec/2_related_work}

\input{sec/3_dataset}
\input{sec/4_method}
\input{sec/5_experiments}
\input{sec/6_limitation}
\input{sec/7_conclusion}

{
    \small
    \bibliographystyle{ieeenat_fullname}
    \bibliography{main}
}

\clearpage
\input{sec/X_suppl}

\end{document}

%% file: preamble.tex
\newcommand{\ours}{URECA dataset\xspace}
\newcommand{\model}{URECA\xspace}
\newcommand{\paragrapht}[1]{\vspace{-10pt}\paragraph{#1}}


%% file: sec/0_abstract.tex

\begin{abstract}

Region-level captioning aims to generate natural language descriptions for specific image regions while highlighting their distinguishing features. However, existing methods struggle to produce unique captions across multi-granularity, limiting their real-world applicability. To address the need for detailed region-level understanding, we introduce \ours, a large-scale dataset tailored for multi-granularity region captioning. Unlike prior datasets that focus primarily on salient objects, \ours ensures a unique and consistent mapping between regions and captions by incorporating a diverse set of objects, parts, and background elements. Central to this is a stage-wise data curation pipeline, where each stage incrementally refines region selection and caption generation. By leveraging Multimodal Large Language Models (MLLMs) at each stage, our pipeline produces distinctive and contextually grounded captions with improved accuracy and semantic diversity.
Building upon this dataset, we present \model, a novel captioning model designed to effectively encode multi-granularity regions. \model maintains essential spatial properties such as position and shape through simple yet impactful modifications to existing MLLMs, enabling fine-grained and semantically rich region descriptions. Our approach introduces \textit{dynamic mask modeling} and a \textit{high-resolution mask encoder} to enhance caption uniqueness. Experiments show that \model achieves state-of-the-art performance on \ours and generalizes well to existing region-level captioning benchmarks.
\vspace{-1.0em}
\end{abstract}

%% file: sec/1_first_revised_intro.tex
\section{Introduction}
\label{sec:intro}
Region-level captioning aims to describe a specific region in natural language while highlighting its distinguishing features compared to other regions. Although previous approaches have shown strong performance in describing target regions, they have struggled to generate \textit{distinguishable captions across multiple granularities}. For real-world applications, it is crucial to describe regions at any level of granularity. 

Despite its importance, generating unique captions across multiple levels of granularity remains underexplored. The primary challenges in multi-granularity captioning include accurately localizing the user-specified region and capturing its unique attributes (e.g., color, relative position, and shape) to clearly distinguish it from surrounding areas~\cite{wang2020uniqueinformativecaptioningimages, liu2019generating, wang2020compare}. Furthermore, generating truly unique regional captions~\cite{Pseudo-RIS, refcap} is particularly challenging, as most existing methods~\cite{DenseCap, kosmos-2, glamm, Yu_2017_CVPR, osprey, controlcap, sca, omg-llava} primarily describe target regions without sufficiently capturing their distinctive attributes regarding their contextual surroundings. Additionally, even though several studies~\cite{internvl-25, llava, improvedllava, arcana} have explored regional captioning, we find that they struggle to produce truly unique and context-aware captions due to limitations in both model design and dataset availability.

Regarding model design, previous approaches~\cite{grit, flexcap, groma, gpt4roi} have explored generating captions on salient regions by providing region coordinates through natural language descriptions~\cite{llava, minigpt}, marking regions directly on the image (e.g., mask contours, ellipses, bounding boxes, triangles, scribbles, points, arrows, and masks)~\cite{vip-llava, yang2023finegrainedvisualprompting, shtedritski2023doesclipknowred}, or utilizing visual RoI features~\cite{regiongpt, omni-rgpt}. Although these methods have demonstrated notable performance, they come with several drawbacks on solving unique multi-granularity region captioning: they may alter the original image, make it difficult to distinguish between colored contours, and fail to capture global relationships within the image~\cite{alpha-clip}, ultimately hindering the model's ability to generate unique captions.

On the other hand, we show that the na\"ive use of existing captioning datasets~\cite{groma, osprey, glamm, visual_genome, refcocog, paco, fullanno, finecaption, use, prima, pascal_voc, partimagenet} is also incompatible with unique multi-granularity captioning. Existing datasets primarily focus on salient regions~\cite{refcocog, glamm, use, prima} or rely on bounding box annotations~\cite{groma, fullanno}, which limits the model’s ability to describe less salient properties. Additionally, most captions in these datasets tend to be overly general~\cite{visual_genome, osprey, paco} (e.g., ``a person standing"), leading to duplicate annotations for multiple regions within the same image. This lack of specificity prevents models from distinguishing similar regions effectively, ultimately hindering their ability to generate faithful descriptions of user-defined regions.

As current datasets have not explored on both generating unique caption and considering multi-granularity, our paper first presents a novel data curation pipeline on generating unique captions on multi-granularity regions. To generate unique captions for regions while considering their hierarchical relationships, our automated data curation pipeline annotates regions in a stage-wise manner using a \textit{mask tree} structure. The mask tree captures hierarchical dependencies by representing subset and superset relationships among regions, as shown in Figure~\ref{fig:main_figure}. By leveraging this structure, our pipeline efficiently generates unique captions for target regions based on their corresponding tree nodes.

Utilizing our data pipeline, we propose \textbf{\ours}, a dataset tailored for unique captioning on multi-granularity regions. Unlike previous captioning datasets~\cite{groma, osprey, glamm, visual_genome, refcocog, paco, fullanno, finecaption, use, prima, pascal_voc} which are solely limited to salient regions within an image and contain simple descriptions, our dataset covers a broader range of objects, parts, and backgrounds each with a unique caption. Each caption in our dataset is uniquely mapped to a single region, ensuring a unique correspondence between regions and captions across various granularities.

Based on our dataset, we then propose a novel model architecture \textbf{\model}, which effectively conditions regions of interest for the captioning model without losing the region's details. To consider multi-granularity regions, it is essential to encode various size of regions, which previous methods often struggled with~\cite{finecaption, regiongpt, sa2va, omg-llava, vip-llava}. To achieve this, we introduce a mask encoder network and dynamic mask modeling approach that extracts mask features while preserving essential properties such as position and shape even on the multi-granularity regions.
\input{tables/dataset_comparison}
To demonstrate the effectiveness of our proposed \textbf{\ours} and \textbf{\model} in generating unique multi-granularity region captions, we further validate our model on a test set that has undergone an additional quality assurance stage to ensure the quality of the annotated captions produced by our automated pipeline. Notably, \model achieves state-of-the-art performance on our \ours test set while also demonstrating strong performance on traditional region-level captioning datasets such as Visual Genome~\cite{visual_genome} and RefCOCOg~\cite{refcocog} datasets. We also show that further fine-tuning existing captioning models on {\ours} enables better multi-granularity captioning.

In summary, we make the following contributions:
\begin{itemize}
\item We introduce a large-scale multi-granularity captioning dataset that ensures unique region-caption mapping by covering a diverse range of objects, parts, and backgrounds beyond salient regions.
\item We present a novel captioning model designed to handle multi-granularity regions through dynamic mask modeling and a high-resolution mask encoder that preserves essential region properties.
\item Extensive experiments demonstrate that our model achieves state-of-the-art performance on our test dataset and shows strong generalization on benchmark datasets, including Visual Genome~\cite{visual_genome} and RefCOCOg~\cite{refcocog}.
\end{itemize}

%% file: tables/dataset_comparison.tex
\begin{table*}[!ht]
    \centering
    \resizebox{1.0\textwidth}{!}{ 
    \begin{tabular}{lccccc}
        \toprule
        Dataset & Simple caption & Dense caption & Region caption & Multi-granularity & Unique caption \\
        \midrule
        RefCOCOg~\cite{refcocog} & \cmark & \xmark & \cmark & \xmark & \xmark \\ 
        Visual Genome~\cite{visual_genome} & \cmark & \xmark & \cmark & \xmark & \xmark \\ 
        PACO~\cite{paco} & \cmark & \xmark & \cmark & \xmark & \xmark \\ 
        Partimagenet~\cite{pascal_voc} & \cmark & \xmark & \cmark & \xmark & \xmark \\
        PRIMA~\cite{prima} & \cmark & \cmark & \xmark & \xmark & \xmark \\
        LLaVA-115K~\cite{llava} & \cmark & \cmark & \xmark & \xmark & \xmark \\
        Arcana~\cite{arcana} & \cmark & \cmark & \cmark & \xmark & \xmark \\ 
        Osprey~\cite{osprey} & \cmark & \cmark & \cmark & \xmark & \xmark \\
        I Dream My Painting~\cite{fanelli2024dream} & \cmark & \cmark & \cmark & \xmark & \xmark \\
        GRIT~\cite{kosmos-2} & \cmark & \cmark & \cmark & \xmark & \xmark \\
        LiSA~\cite{lisa} & \cmark & \cmark & \cmark & \xmark & \xmark \\
        USE~\cite{use} & \cmark & \cmark & \xmark & \cmark & \xmark \\
        SegCAP~\cite{mgllm} & \cmark & \cmark & \cmark & \cmark & \xmark \\ 
        GranD~\cite{glamm} & \cmark & \cmark & \cmark & \cmark & \xmark \\ 
        \midrule
        \textbf{\ours (Ours)} & \cmark & \cmark & \cmark & \cmark & \cmark \\
        \bottomrule
    \end{tabular}
    }
    \caption{\textbf{Statistical comparison of previous captioning datasets and \ours in region-level captioning.} The comparison covers different types of captions, including simple captions (\textit{e.g.}, \cite{paco, partimagenet}), dense captions (\textit{e.g.}, \cite{prima, llava}), region captions (\textit{e.g.}, \cite{refcocog, visual_genome, arcana, osprey, grit, fanelli2024dream, lisa}), and multi-granularity captions (\textit{e.g.}, \cite{use, mgllm, glamm}). While these datasets provide varying levels of detail, \ours is the only dataset that offers distinctive dense captions and handles multi-granularity regions effectively.}
    \vspace{-10pt}

    \label{tab:dataset_comparison}
\end{table*}

%% file: sec/2_related_work.tex
\vspace{-5pt}
\section{Related Work}
\label{sec:related_work}

\paragraph{Multi-modal large language model.}
Large Language Models (LLMs) have demonstrated pioneering performance in instruction following capabilities, integrating diverse knowledge from extensive datasets, and performing complex reasoning tasks. However, a significant limitation of LLMs is their reliance solely on natural language inputs. To address this, LLaVA~\cite{llava} was the first to explore the integration of image and text modalities by representing visual features as visual tokens.  Building upon this, models such as Flamingo~\cite{flamingo} and BLIP-2~\cite{blip-2} have further advanced Multimodal Large Language Models (MLLMs) by incorporating powerful visual backbones. These models effectively bridge the two modalities and have shown strong performance in tasks like image captioning and visual question answering. Building on these advancements, recent efforts
have aimed to extend these models
to handle more complex tasks, including reasoning over segmentation~\cite{lisa, ren2024pixellm}, optical character recognition~\cite{qwen2-vl, dong2024internlmxcomposer}, and grounding~\cite{flickr, glamm, use, mgllm, grain}.

\paragrapht{Region-level vision language model.}
Although MLLMs have demonstrated impressive image understanding capabilities, generating captions for specified regions remains a challenging task. LLaVA~\cite{llava} and MiniGPT-2~\cite{minigpt} have explored conditioning given regions by translating bounding box coordinates into natural language. However, these models heavily rely on the MLLMs’ ability to interpret bounding box coordinates accurately. Other approaches~\cite{vip-llava, yang2023finegrainedvisualprompting, shtedritski2023doesclipknowred} have attempted to overlay regions directly onto the image. While this method is straightforward to implement, it alters the original image, making it difficult for MLLMs to reference the unmodified content.
To address this issue without modifying the original image, some methods have explored directly modeling the coordinates of the regions or feature pooling conditioned on bounding boxes~\cite{grit, flexcap, groma, gpt4roi}. Although pooling features from the bounding box has improved performance, these approaches often struggle to accurately capture user intent, particularly when objects overlap. Mask-based feature pooling~\cite{regiongpt, omni-rgpt} provides more precise localization information by avoiding ambiguous bounding box indications. However, it is typically performed on low-resolution image features and excessively aggregates information, leading to the loss of fine-grained details such as shape and boundaries. In extreme cases, small-region masks in high-resolution images may disappear entirely during this process, resulting in the loss of meaningful features. 

Moreover, none of the prior works have effectively addressed the challenge of generating captions that precisely localize user-intended regions while capturing their unique attributes at any granularity. This is primarily due to the lack of a suitable dataset and the absence of architectures designed for this task. To bridge this gap, we propose an automated data generation pipeline that ensures the inclusion of unique captions while considering multi-granularity regions. Additionally, our architecture effectively handles such multi-granularity regions, preserving their original attributes and capturing global relationships among regions.



%% file: sec/3_dataset.tex
\section{URECA Dataset} 

\input{figs/main_framework}
Previous research has made significant progress in generating dense region captions; however, approaches focusing on multi-granularity regions remain scarce. When considering the granularity of regions, distinguishing their unique attributes becomes crucial~\cite{refcap, wang2020uniqueinformativecaptioningimages, liu2019generating, wang2020compare}, as visually similar regions frequently appear within an image. Existing approaches have struggled to generate truly unique captions for regions, often producing generic descriptions despite clear visual differences.

This tendency to generate generic captions contradicts human perception, as humans naturally recognize and describe regions based on distinctive attributes like color, position, and shape. However, existing captioning datasets often lack such specificity, and training models on such generic captions that do not emphasize regional uniqueness can contribute to the \textit{mode collapse} problem~\cite{wang2020uniqueinformativecaptioningimages}, where models fail to generate diverse and informative captions. 


To address this lack of specificity in existing datasets, we propose \ours, a dataset designed to enhance models' ability to generate unique captions for given multi-granularity regions. Our dataset is generated through an automated data pipeline that creates and verifies captions in a stage-wise manner. Specifically, we build our dataset using the publicly available SA-1B dataset, which offers high-resolution images and multi-granularity regions. To further ensure caption quality in the test set, we incorporate a verification step using GPT-4o~\cite{gpt4o} as part of the pipeline.

\paragrapht{Data annotation pipeline.}
To generate unique captions that effectively capture multi-granularity, it is crucial to consider both target and non-target regions. Captions that focus solely on the target region often become overly localized and repetitive, making it difficult to distinguish between similar regions. To address this, we structure hierarchical relationships between regions, ensuring that captions incorporate broader contextual information.

At the core of our approach is a mask tree, constructed based on Intersection-over-Union (IoU). This hierarchical structure organizes regions into subset-superset relationships, allowing us to systematically capture dependencies between different regions. This hierarchical structure enables a comprehensive understanding of region dependencies at both global and local levels, ensuring the generation of unique captions. 

This process follows a structured sequence of four stages, as illustrated in Figure~\ref{fig:main_figure}:
\begin{enumerate}[leftmargin=*] 
    \item \textbf{Mask tree generation.} We first construct a mask tree to represent the hierarchical relationships among masks in an image. By comparing the IoU between masks, we can determine their relationships (i.e., superset or subset) within the hierarchy.
    
    \item \textbf{Top-down generation.} To ensure that contextual information is effectively incorporated into each node’s caption, we generate captions in a top-down manner. In this process, each node refers to its parent node to maintain hierarchical consistency. Specifically, we generate short captions using our annotation MLLM, InternVL2.5-38B~\cite{internvl-25}, for each node by referring captions from the parent node and two types of images that represent the target region: a cropped image of the target region with non-target areas blurred based on the mask~\cite{blur}, and a cropped image of the parent region, where the target region is contoured while non-target areas within the parent region are blurred.
    
    \item \textbf{Bottom-up generation.} To ensure that parent nodes have unique captions incorporating relevant details from their child nodes while maintaining contextual coherence, we generate captions in a bottom-up manner. In this process, the parent node refers to its children's captions to generate a more informative and unique caption. Specifically, we aggregate the captions of all child nodes and use our annotation MLLM to generate a refined caption based on the aggregated captions, the parent node’s short caption, and an image where the target region is contoured within the full image to preserve its spatial context.

    \item \textbf{Uniqueness refinement.} To further ensure visually similar regions have distinguishable captions, we introduce a uniqueness refinement process based on image feature similarity using DINOv2~\cite{dinov2}. In this stage, similar-looking regions are identified using image features and marked in the image with contours and indexed bounding boxes~\cite{yang2023setofmark}. Our annotation MLLM then generates a unique caption by explicitly differentiating the target region from other visually similar regions.
\end{enumerate}

\vspace{-5pt}

\paragrapht{Data statistics.}
We conducted a statistical comparison between previous captioning datasets and \ours. Table~\ref{tab:dataset_comparison} highlights their capabilities in region-level captioning. Simple caption refers to datasets~\cite{paco, partimagenet} that provide basic descriptions, often incorporating object classes in the captions. Dense caption represents datasets~\cite{prima, llava} that include multiple attributes, offering more detailed descriptions of the region. Additionally, datasets~\cite{refcocog, visual_genome, arcana, osprey, grit, fanelli2024dream, lisa} where captions are explicitly aligned with specific regions fall under the region caption category. As multi-granularity captioning becomes increasingly relevant for real-world applications, recent datasets~\cite{use, mgllm, glamm} have started to incorporate this aspect. 
However, none of the existing datasets fully capture all these aspects with captions that describe distinctive attributes of the region while maintaining multi-granularity.
Among them, \ours stands out as a unique dataset providing distinct dense captions while effectively handling multi-granularity regions.

\vspace{-5pt}

\paragrapht{Evaluation set.}
To ensure the quality of the test dataset when evaluating unique captioning on multi-granularity regions, we additionally implemented a verification stage during the test set generation process. As state-of-the-art MLLMs have demonstrated performance comparable to human annotators’ preferences~\cite{lee2024prometheus_vision, xiong2024llava_critic, ge2023mllm_bench}, we utilized GPT\footnote{gpt-4o-mini-2024-07-18}, which is widely adopted to simulate human annotators for data generation tasks. Further details about the dataset pipeline can be found in Appendix.

%% file: figs/main_framework.tex
\begin{figure*}[!ht]
    \centering
    \includegraphics[width=\linewidth]{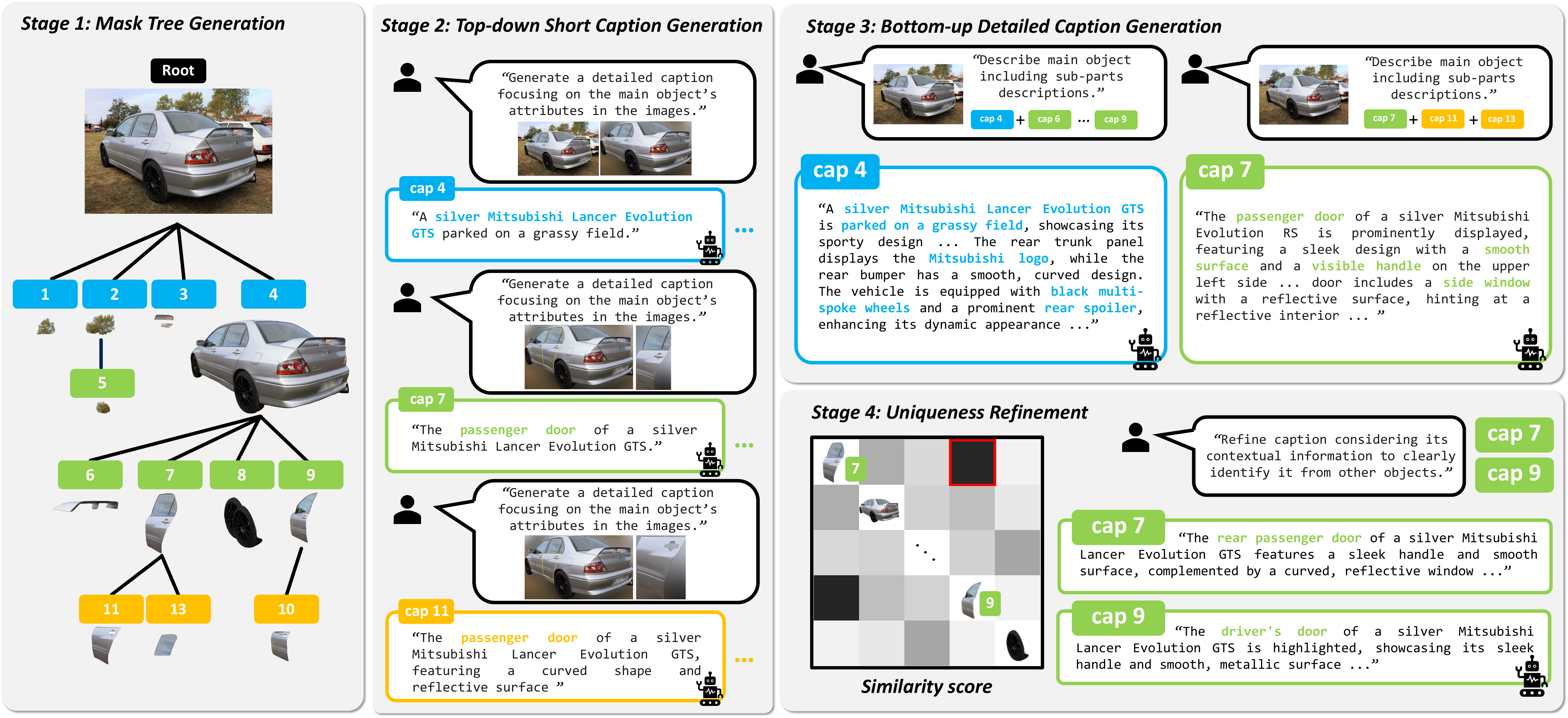}
    \caption{\textbf{Automated data curation pipeline of \ours.} Our pipeline consists of four key stages to generate unique captions for multi-granularity regions. In Stage 1, we construct a mask tree that captures hierarchical relationships between regions. Stage 2 generates short captions based on the parent node. Stage 3 aggregates captions from child nodes, and Stage 4 ensures that each node is assigned a unique caption. Best viewed in zoomed-in.}
    \vspace{-10pt}
    \label{fig:main_figure}
\end{figure*}

%% file: sec/4_method.tex
\section{URECA}


\input{figs/model_architecture}
The overall architecture of \model is illustrated in Figure~\ref{fig:model_arch}. It builds upon the LLaVA~\cite{llava} architecture, which treats visual features as tokens. Inspired by this approach, we adopt a similar strategy by representing mask features as tokens. To achieve this, we integrate a mask encoder that localizes the target region while preserving essential mask details such as size, position, and shape.


\subsection{Mask Encoder}
To capture the distinctive features of a target region, it is essential to leverage both local information and the global context of the image. A representation of the region should function as a localizer rather than a constraint on the region. 
Additionally, the representation of the region should encode information such as size, position, and shape without ambiguity or loss. Masks inherently provide this capability, whereas previous methods such as contours, bounding boxes, and text coordinates do not. Therefore, encoding mask information without loss is crucial for generating dense and distinctive captions. 


To achieve this, we introduce a mask encoder that exclusively encodes the mask without altering the original image, thereby preserving the mask's unique attributes. Our mask encoder transforms the binary mask into a sequence of tokens through multiple convolutional layers, resulting in a set of mask tokens. These mask tokens are then integrated with image tokens within the MLLM, enabling the mask to function effectively as a localizer while maintaining precise region-specific information. More precisely, our mask encoder performs as:
\begin{equation}
\mathbf{F} = \phi(\mathbf{M}) \in \mathbb{R}^{N \times D},
\end{equation}
where \(\mathbf{M} \in \{0, 1\}^{H \times W}\) denote the input binary segmentation mask, and \(H\) and \(W\) represent the height and width of the mask, respectively. The encoding process is performed by a mask encoder, represented by the function \(\phi(\cdot)\). Our mask encoder maps the binary mask \(\mathbf{M}\) to a feature representation \(\mathbf{F} \in \mathbb{R}^{N \times D}\), where \(N\) denotes the number of spatial tokens, and \(D\) is the feature embedding dimension.

Unlike traditional mask feature pooling, our mask encoder tokenizes the mask without aggregation, preserving fine-grained spatial details and capturing multi-granularity regions. The resulting mask tokens contain both local and global details of the mask, enabling our model to accurately locate regions and generate clear, unique captions.

\subsection{Dynamic Mask Modeling}
Na\"ively using an encoder that receives fixed-size inputs requires mask resizing, leading to the loss of fine-grained region details. However, preserving these details is particularly important, as multi-granularity captioning relies on masks of diverse scales. Therefore, generating mask tokens directly from high-resolution inputs is essential.

Thanks to the success of MLLMs that accept dynamic-length inputs, we leverage this capability in our mask modeling. To provide a precise view during mask modeling and alleviate the challenges caused by extensive resizing, we propose dynamic masking which split the original high-resolution mask into multiple sub-masks. This approach allows the length of mask tokens to be dynamically adjusted based on resolution, ensuring a more accurate and flexible representation.

Specifically, before passing the mask through the encoder, we first apply a dynamic masking process to split the original mask \(\mathbf{M} \in \{0, 1\}^{H \times W}\) into multiple sub-masks \(\mathbf{M}_{\text{split}}\) as follows:
\begin{equation}
\mathbf{M}_{\text{split}} = \text{Split}(\mathbf{M}) \in \{0, 1\}^{N_s \times H' \times W'}.
\end{equation}
These sub-masks are obtained by dividing the original mask into smaller regions, each having a size of \(H' \times W'\). The number of sub-masks \(N_s\) depends on the pre-defined splitting strategy. This process ensures that the mask encoder receives high-resolution and localizes information while preserving the global context of the original mask. The sub-masks \(\mathbf{M}_{\text{split}}\) are then passed through the mask encoder, resulting in the tokenized feature representation \(\mathbf{F}_{\text{split}} \in \mathbb{R}^{N_s \times D}\), where \(D\) is the feature embedding dimension. This dynamic masking step allows for finer localization of regions and helps capture multi-granularity features before encoding them into a compact representation.


\label{sec:method}

%% file: figs/model_architecture.tex
\begin{figure*}[!ht]
    \centering
    \vspace{-5pt}
    \includegraphics[width=\linewidth]{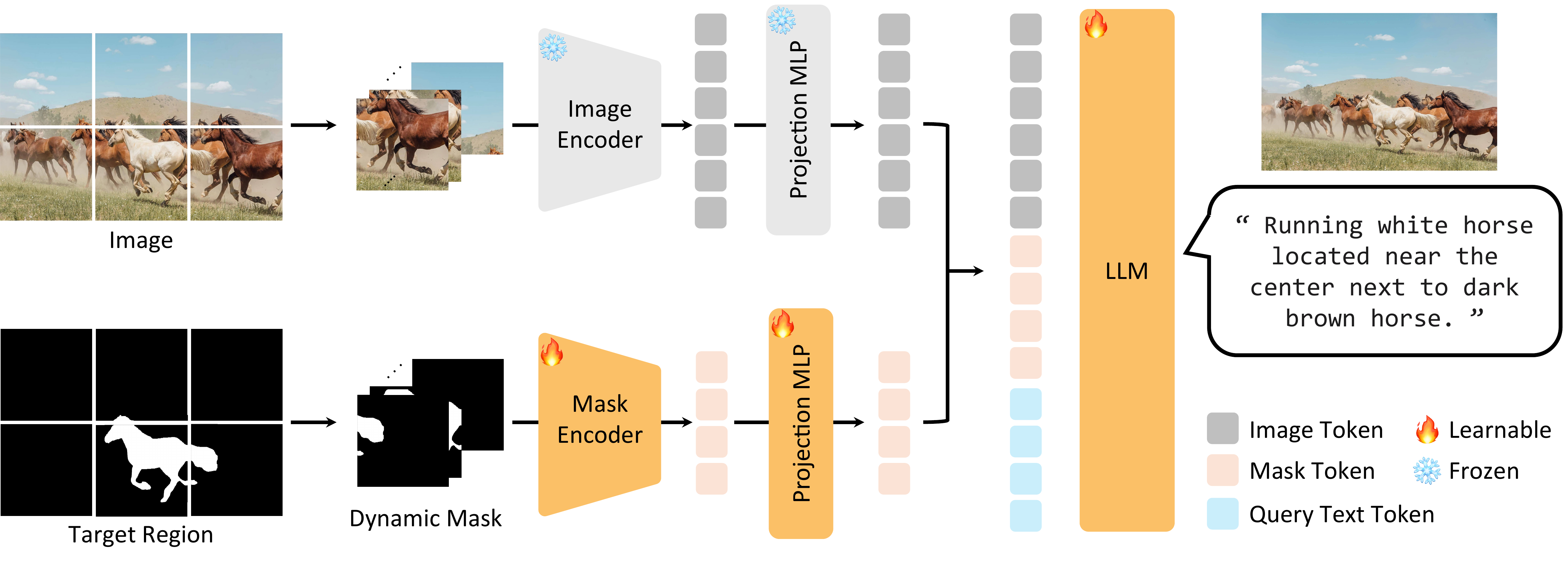}
    \vspace{-10pt}
    \caption{\textbf{\model architecture.} URECA enables users to generate unique captions that describe distinctive attributes of any region. The mask encoder effectively encodes multi-granularity regions while preserving their identity. The mask token serves as a localizer, guiding the LLM to generate region-specific captions based on the image and query token.}
    \vspace{-10pt}
    \label{fig:model_arch}
\end{figure*}

%% file: sec/5_experiments.tex
\section{Experiments}
\label{sec:experiments}
\input{tables/mug-cap_table}



\subsection{Quantitative Results}

\input{tables/zeroshot_implementation}
We report the performance of \model on \ours as well as previous benchmark datasets~\cite{visual_genome, refcocog}. All results are evaluated using an 8B language model trained exclusively on the \ours.

\paragrapht{Unique multi-granularity region captioning.}
In Table~\ref{tab:main_table}, we present the performance comparison on \ours, a dataset specifically designed to evaluate unique multi-granularity region captions, alongside previous methods. To demonstrate the effectiveness of our approach, we implemented a baseline by running a na\"ive MLLM~\cite{internvl-25} on \ours. ``None" refers to providing the MLLM with only the image, without any explicit region marking. ``Contour" refers to marking regions within the image, and ``Crop" involves providing the MLLM with a cropped view of the target region. The results indicate that conditioning the MLLM solely on the image or natural language fails to localize regions effectively and generate unique captions.  

While previous region-level captioning models~\cite{groma, sca, kosmos-2, gpt4roi, omg-llava, vip-llava} have demonstrated improved performance in generating unique captions when trained on \ours, they lag behind \model either because they struggle to localize multi-granularity regions, alter the original image, or overly constrain the target region without considering the global context.

This underscores that fine-tuning existing captioning models on the \ours enhances their ability to handle multi-granularity captioning. However, \model surpasses these approaches by not only generating unique captions across an image but also effectively capturing multi-granularity regions, demonstrating its capability to accurately represent regional information.

\paragrapht{Region-level captioning.}

In Table~\ref{tab:zeroshot}, we present the zero-shot performance of \model on RefCOCOg~\cite{refcocog} and Visual Genome~\cite{visual_genome}. On RefCOCOg, \model demonstrated competitive performance, while on Visual Genome, it achieved state-of-the-art results compared to previous approaches.

Notably, unlike prior methods, \model achieves these results without using the benchmarks’ training sets, highlighting the strong generalization ability of \ours. This suggests that \ours covers diverse region granularities with well-aligned captions, enabling better regional understanding. By effectively learning from a dataset with varying granularities, \model effectively localizes and generates captions across different scales, making it highly adaptable to region-level captioning even on the zero-shot tasks.

\subsection{Qualitative Results}
\input{figs/qualitatives}
Figure~\ref{fig:qual_results} shows the qualitative results comparing \model with other methods~\cite{kosmos-2, omg-llava, vip-llava}.Unlike the comparison models, which struggle to localize regions or generate captions that capture their distinctive attributes, \model produces unique captions for each multi-granularity region. Additional qualitative results are provided in the Appendix.


\subsection{Ablation Studies}
\paragraph{Effectiveness of mask encoder.}
\input{tables/ablations/dynamic_mask}
To evaluate the effectiveness of our proposed methods, we conduct an ablation study by separately implementing each component and assessing their impact on model performance. As presented in Table~\ref{tab:ablation-methods}, the baseline MLLM without conditioning performs poorly. Incorporating our mask encoder, which effectively encodes the target region while preserving its identity, significantly enhances the model's ability to localize regions and generate more descriptive captions. Furthermore, employing our dynamic masking strategy, which divides the original resolution into smaller sub-images, enables the mask encoder to capture finer details of target regions, further improving performance.

\vspace{-2pt}
\paragrapht{MLLM size.}
\input{tables/ablations/model_size}
It is well established that performance improves with larger foundation models~\cite{li2024lmms, internvl-25, zhang2022opt, qwen2_5}, as their knowledge capacity scales with model size. Our \model follows this trend, achieving better performance as its size increases, as shown in Table~\ref{tab:ablation-model-size}. While the 1B model records the lowest performance, the largest model (8B) achieves the highest.

\paragrapht{Mask token length.}
\input{tables/ablations/mask_token_length}
We demonstrated that our mask encoder effectively captures regions while preserving their identity. To analyze the impact of the number of tokens generated by the mask encoder, we conduct an ablation study, as shown in Table~\ref{tab:num_mask_token}. 
We investigate the effect of increasing the number of mask tokens. As the number of tokens increases, the representation becomes more detailed, allowing for finer details to be captured, particularly in smaller regions.


%% file: tables/mug-cap_table.tex
\begin{table*}[!ht]
\centering
\resizebox{0.94\textwidth}{!}{ 
\begin{tabular}{l|ccccccc}
\toprule
Models    & BLEU@1 & BLEU@2 & BLEU@3 & BLEU@4 & ROUGE & METEOR & BERTScore \\
\midrule
None      & 17.06 & 7.63 & 3.14 & 1.20 & 17.86 & 27.72 & 62.68 \\
Contour   & 17.10 & 7.13 & 2.63 & 1.01 & 19.95 & 25.49 & 63.29 \\
Crop      & 18.43 & 7.53 & 2.45 & 0.85 & 19.73 & 26.45 & 63.63 \\
\midrule
SCA~\cite{sca}        & 22.76 & 13.58 & 6.97 & 3.88 & 30.76 & 24.87 & 70.67 \\
KOSMOS-2~\cite{kosmos-2} & 30.31 & 18.12 & 9.96 & 5.55 & 34.19 & 32.94 & 72.64 \\
OMG-LLaVA~\cite{omg-llava} & 34.01 & 21.88 & 13.51 & 8.46 & 38.14 & 37.29 & 74.68 \\
ViP-LLaVA~\cite{vip-llava} & 34.17 & 22.07 & 13.96 & 9.00 & 38.17 & 37.68 & 74.62 \\
\midrule
\textbf{URECA (Ours)}       & \textbf{36.56} & \textbf{23.84} & \textbf{15.42} & \textbf{9.98} & \textbf{38.95} & \textbf{41.25} & \textbf{75.11} \\
\bottomrule
\end{tabular}
}
\vspace{-5pt}
\caption{\textbf{Performance comparison of \model with baseline methods and previous models on various evaluation metrics, including BLEU~\cite{bleu}, ROUGE~\cite{rouge}, METEOR~\cite{meteor}, and BERTScore~\cite{bert-score}.} The results show that \model outperforms other methods across all metrics on URECA testset, demonstrating its superior ability to generate unique captions for multi-granularity regions. Note that comparison methods are all trained on \ours.}
\vspace{-5pt}
\label{tab:main_table}
\end{table*}

%% file: tables/zeroshot_implementation.tex
\begin{table}[t]
\centering
\resizebox{\columnwidth}{!}{ 
\begin{tabular}{l|cc}
\toprule
Models & RefCOCOg & Visual Genome \\
\midrule
ControlMLLM~\cite{controlmllm}           &    14.0     &     -     \\
Kosmos-2~\cite{kosmos-2} &     14.1    &     -     \\
GRiT~\cite{grit}        &    15.2     &     17.1      \\
SLR~\cite{slr}                    &   15.9      &      -    \\
GLaMM~\cite{glamm}      &    15.7     &      17.0      \\
OMG-LLaVA~\cite{omg-llava}  &    15.3     &    -     \\
ViP-LLaVA~\cite{vip-llava}  &    16.6     &    -      \\
Groma~\cite{groma}      &    16.8     &    16.8      \\
RegionGPT~\cite{regiongpt}  &    16.9     &    17.0      \\
Omni-RGPT~\cite{omni-rgpt}  &    \textbf{17.0}     &     17.0     \\
\midrule
\textbf{URECA~(Zero-Shot)}                   &    16.1     &     \textbf{18.4}     \\
\midrule
\end{tabular}
}
\vspace{-5pt}
\caption{\textbf{Quantitative results on region-level captioning task.} Performance comparison on the METEOR~\cite{meteor} for the RefCOCOg~\cite{refcocog} and Visual Genome~\cite{visual_genome} datasets. (Zero-Shot) refers to zero-shot transfer.}
\vspace{-15pt}
\label{tab:zeroshot}
\end{table}

%% file: figs/qualitatives.tex
\begin{figure*}[!ht]
    \centering
    \vspace{-5pt}
    \includegraphics[width=\linewidth]{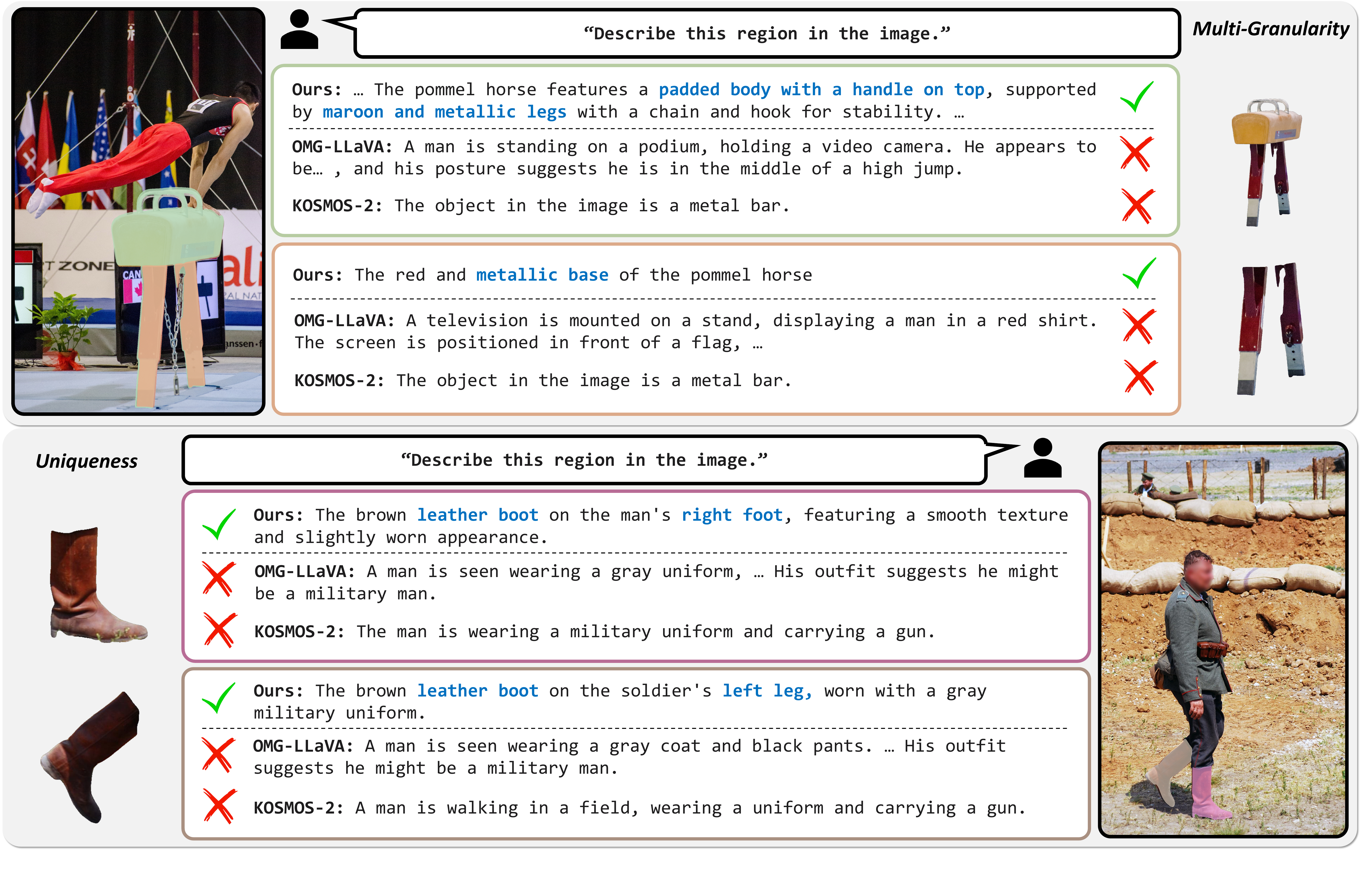}
    \vspace{-25pt}
    \caption{\textbf{Qualitative results of the \model and comparison models~\cite{kosmos-2, omg-llava}.} Our model generates unique caption conditioned on multi-granularity regions.}
    \vspace{-10pt}
    \label{fig:qual_results}
\end{figure*}

%% file: tables/ablations/dynamic_mask.tex


\begin{table}[t]
\centering
\resizebox{0.90\columnwidth}{!}{ 
\begin{tabular}{l|ccc}
    \toprule
    \textbf{Method} & ROUGE & METEOR & BERTScore \\
    \midrule
    Baseline        & 17.86  & 27.72  & 62.68  \\
    + Mask Encoder  & 38.46  & 40.72  & 74.73  \\
    + Dynamic Mask  & 38.95  & 41.25  & 75.11  \\
    \bottomrule
\end{tabular}
}
\vspace{-5pt}
\caption{\textbf{Ablation study of our proposed methods on \ours.}}
\vspace{-10pt}
\label{tab:ablation-methods}
\end{table}

%% file: tables/ablations/model_size.tex


\begin{table}[t]
\centering
\resizebox{0.9\columnwidth}{!}{ 
\begin{tabular}{c|ccc}
    \toprule
    \textbf{Model Size} & ROUGE & METEOR & BERTScore \\
    \midrule
    1B  & 32.00  & 33.99  & 71.77  \\
    2B  & 36.64  & 39.00  & 73.92  \\
    4B  & 36.58      &  38.75     & 73.97      \\
    8B  & 38.95  & 41.25  & 75.11  \\
    \bottomrule
\end{tabular}
}
\vspace{-5pt}
\caption{\textbf{Ablation study on model size.}}
\vspace{-5pt}
\label{tab:ablation-model-size}
\end{table}

%% file: tables/ablations/mask_token_length.tex


\begin{table}[t]
\centering
\resizebox{0.9\columnwidth}{!}{ 
\begin{tabular}{c|ccc}
    \toprule
    \textbf{Token Length} & ROUGE & METEOR & BERTScore \\
    \midrule
    4  & 35.44 & 38.01 & 73.51 \\ 
    8  & 37.06 & 38.50 & 74.21 \\
    16 & 38.95 & 41.25 & 75.11 \\
    \bottomrule
\end{tabular}
}
\vspace{-5pt}
\caption{\textbf{Ablation study on mask token length.}}
\vspace{-15pt}
\label{tab:num_mask_token}
\end{table}

%% file: sec/6_limitation.tex


%% file: sec/7_conclusion.tex
\vspace{-5pt}
\section{Conclusion}
\label{sec:conclusion}

We present \ours, a regional captioning dataset that includes multi-granularity regions. Our primary objective is to annotate regions with unique captions that exclusively describe the target region. To achieve this, we propose an automated data pipeline that generates distinctive captions using a mask tree, which captures the hierarchical relationships between regions. To ensure high-quality evaluation, we introduce a verification stage to validate the test set. Furthermore, we introduce \model, which encodes masked regions while effectively preserving their identity. To retain finer details, we propose dynamic masking, leveraging the LLM’s flexible input length to encode masks even in high-resolution views. Our \model achieved state-of-the-art performance on \ours and demonstrates strong zero-shot captioning capabilities.

%% file: sec/X_suppl.tex
\clearpage
\maketitlesupplementary
\renewcommand{\thesection}{\Alph{section}}
\setcounter{section}{0} 
\renewcommand{\thetable}{\Alph{table}}
\setcounter{table}{0}
\renewcommand{\thefigure}{\Alph{figure}}
\setcounter{figure}{0}

\section{Implementation Details}
We leverage InternVL-2.5~\cite{internvl-25} along with our mask encoder, which consists of convolutional layers followed by a two-layer MLP as the projection layer for mask tokens. For our experiments, we set the mask token length to 8. The input to the mask encoder is resized to 448×448, and the dimension of the mask tokens matches the feature dimension of the MLLM.  

We train our model on four Tesla A100 GPUs (40GB) using LoRA~\cite{LoRA}. Specifically, training is conducted in two stages: first, we train the mask encoder and projection layer, followed by LoRA fine-tuning of the MLLM. We use a batch size of 16 for LoRA tuning.  

For evaluation, we adopt standard metrics used in previous studies, including BLEU~\cite{bleu}, ROUGE~\cite{rouge}, METEOR~\cite{meteor}, and BERTScore~\cite{bert-score}.

\section{Limitations}
While our mask encoder effectively encodes multi-granularity regions without losing details, localizing the region in a sequential manner may occasionally cause the MLLM to misidentify the target region. Since we do not explicitly constrain target regions using image features or direct markers, the localization signal provided to the MLLM may be weaker compared to previous methods. Enhancing region encoding by incorporating both the mask and additional image features, rather than relying solely on sequential conditioning, could improve the MLLM's ability to accurately localize the target region.

\section{More Qualitative Results}
We visualize more qualitative results of \model with previous apporaches~\cite{vip-llava, omg-llava} in Figure~\ref{fig:supple/pred_1}.

\section{Dataset Visualization}
We provide visual examples of our dataset to illustrate its diversity and complexity. Figure~\ref{fig:supple/data_1} showcases representative samples, highlighting key variations in object appearance, background context, and challenging scenarios. For optimal viewing, we recommend zooming in and viewing the figures in color to better observe fine details.

\section{Data Pipeline}
\label{sup:data_pipeline}
To generate unique regional captions with multi-granularity, we propose a structured four-stage process:

\paragraph{Stage 1: Mask Tree Construction.} We first build a mask tree for each image using masks from the SA-1B dataset~\cite{sam}. Intersection over Union (IoU) between masks is computed to determine containment relationships. Each tree has a root node representing the entire image, with subsequent nodes structured hierarchically based on these containment relationships.
\paragraph{Stage 2: Top-Down Caption Generation.} In this stage, we identify primary nodes directly under the root node, termed \textit{main objects}, whose depth exceeds a predefined threshold. Short captions are then hierarchically generated from these main objects downward through descendant nodes. Each node creates concise captions using contextual information from parent and sibling nodes to maintain coherence and uniqueness. Specific prompts used in this step are detailed in Table~\ref{tab:supp_prompt_top_down}.
\paragraph{Stage 3: Bottom-Up Caption Refinement.} Short captions generated in Stage 2 are expanded into detailed descriptions. Each node enriches its caption by incorporating information from child nodes, ensuring hierarchical consistency and comprehensive context. Prompts for this refinement stage are provided in Table~\ref{tab:supp_prompt_bottom_up}.
\paragraph{Stage 4: Uniqueness Refinement.} Finally, captions are refined by evaluating visual similarity between regions using DINO v2~\cite{dinov2}. Regions with high visual similarity have their captions adjusted by emphasizing distinguishing features, maintaining semantic relevance and uniqueness. Prompts for uniqueness refinement are described in Table~\ref{tab:supp_prompt_unique}.

Through these stages, we systematically generate multi-granularity captions that accurately describe each region with clarity, context, and uniqueness in an automated manner.

\section{Discussion}
Evaluating unique caption generation for regional captioning tasks using traditional metrics such as BLEU~\cite{bleu}, METEOR~\cite{meteor}, ROUGE~\cite{rouge}, and CIDEr~\cite{cider} presents inherent limitations. These metrics primarily assess similarity to reference captions based on n-gram overlap, without distinguishing between essential and non-essential words. However, in unique captioning, it is crucial to generate descriptions that highlight distinctive attributes, ensuring that the caption effectively differentiates the target region from others. Existing evaluation methods treat all words equally, failing to account for the importance of discriminative terms. As a result, captions that successfully emphasize key distinguishing features may not receive high scores if their phrasing deviates from reference texts, even if they better serve the task's objective. This limitation suggests the need for alternative evaluation approaches that better capture the quality and distinctiveness of unique captions.

\input{figs/supple/pred_1}
\input{figs/supple/data_1}
\input{tables/supplementary/prompt_top_down}
\input{tables/supplementary/prompt_bottom_up}
\input{tables/supplementary/prompt_unique}



%% file: figs/supple/pred_1.tex
\begin{figure*}[!ht]
    \centering
    \includegraphics[width=\linewidth]{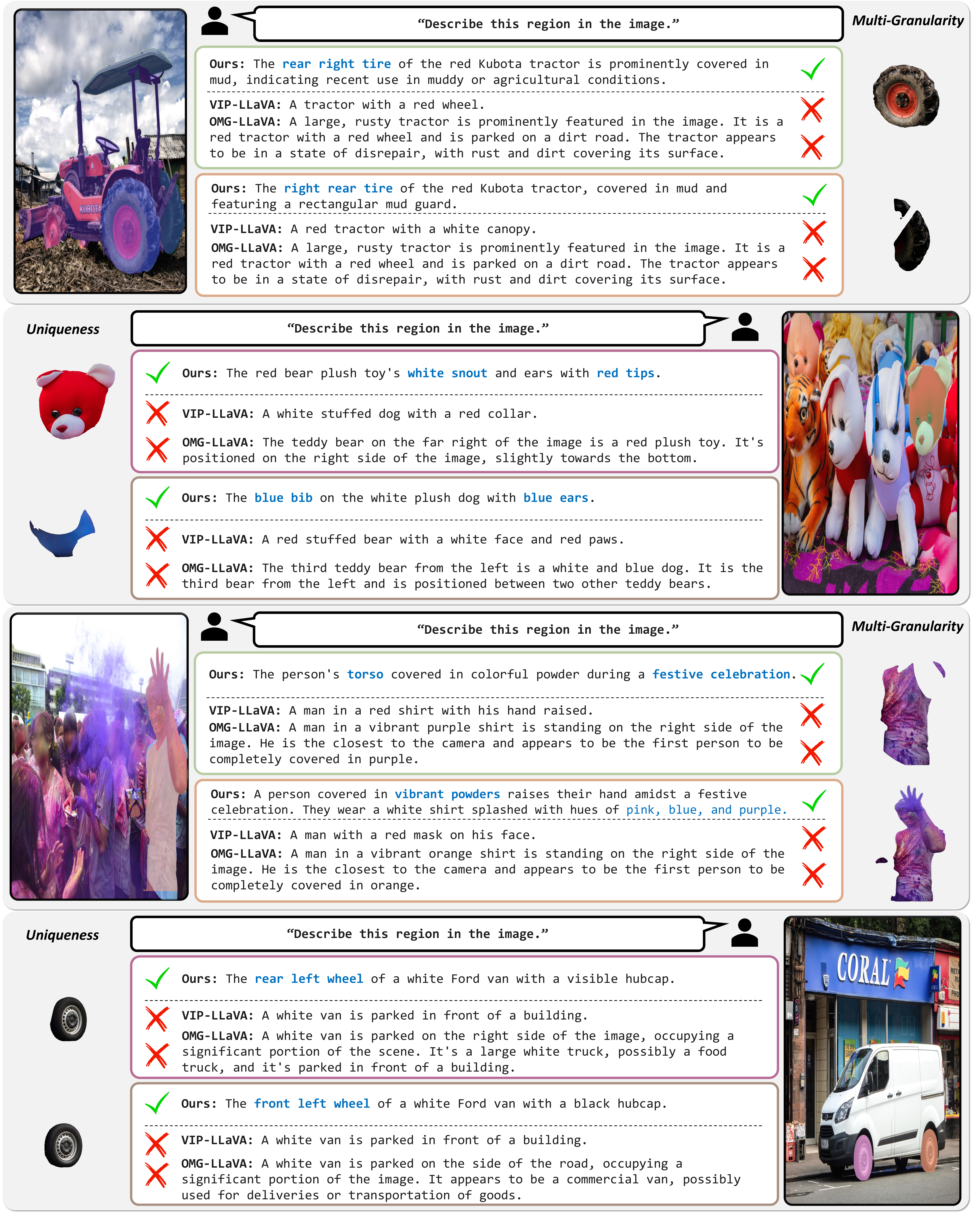}
    \vspace{-10pt}
    \caption{\textbf{Qualitative results of the \model and comparison models~\cite{vip-llava, omg-llava}.} Our model generates unique caption conditioned on multi-granularity regions.}
    \vspace{-10pt}
    \label{fig:supple/pred_1}
\end{figure*}

%% file: figs/supple/data_1.tex
\begin{figure*}[!ht]
    \centering
    \includegraphics[width=\textwidth]{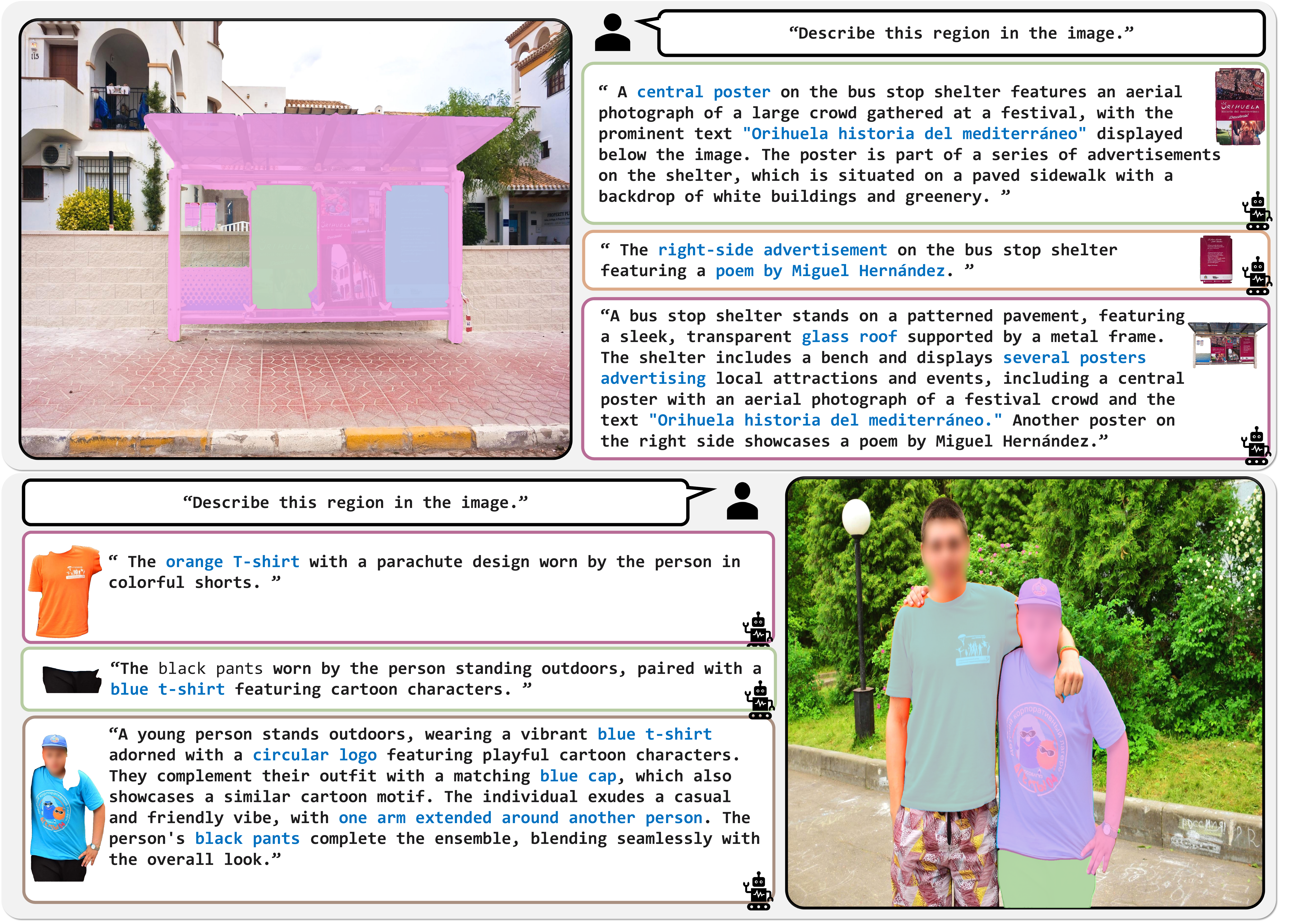}
    \vspace{-10pt}
    \caption{\textbf{Example data generated by our data curation pipeline.} }
    \vspace{-10pt}
    \label{fig:supple/data_1}
\end{figure*}

%% file: tables/supplementary/prompt_top_down.tex
\definecolor{codebg}{rgb}{0.95, 0.95, 0.95}  
\lstdefinestyle{mystyle}{
    backgroundcolor=\color{codebg},   
    basicstyle=\ttfamily\small,       
    frame=single,                    
    breaklines=true,       
    breakautoindent=false,
    captionpos=b,                      
    keywordstyle=\bfseries,            
    commentstyle=\color{gray},         
    numbersep=5pt,                    
    xleftmargin=5pt, xrightmargin=5pt,
    prebreak={},  
    postbreak={}, 
}

\begin{table*}[t]
    \centering
    \begin{minipage}{\textwidth}
        \begin{lstlisting}[style=mystyle]
<task>
    You are a detailed caption generator tasked with describing the main object in images.
    Your goal is to create a simple phrase that accurately represents the main object
    while avoiding hallucination.
</task>
<objectives>
    1. The main object is a subpart of a larger object; therefore, the main object alone
       may provide insufficient information.
    2. The primary focus of the caption must be on the main object while also considering
       its positional relationship or functional connection with the larger object.
    3. The primary focus of the caption must be on the main object, emphasizing attributes
       like color, texture, shape, and action if visible.
    4. The background is blurred to emphasize the main object. Focus solely on describing
       the main object in detail without mentioning the blurred background.
    5. The caption should be distinguishable from other subparts of the same larger
       object so that the region can be identified solely by looking at the caption.
       Therefore, the caption should incorporate positions or attributes that are unique
       to the main object.
    6. Creating a unique caption is important, but the most critical aspect is accuracy.
       Do not add unnecessary information solely for the sake of uniqueness.
</objectives>
<inputDetails>
    1. Image-1 highlights the main object with a yellow contour to illustrate its
       relationship with the larger object.
    2. Image-2 shows the main object cropped from the larger object.
    3. A description of the larger object will be provided in the prompt to help
       identify the main object.
    4. Descriptions of other subparts of the same larger object will also be provided.
       The caption for the main object must be clearly distinguishable from the
       descriptions of these subparts.
</inputDetails>
<descriptionOfLargerObject>
    "Description from the parent object"
</descriptionOfLargerObject>
<descriptionOfSubparts>
    "Descriptions from objects on the same level, if present."
</descriptionOfSubparts>
<outputFormat>
    1. Provide a simple phrase focusing on the main object while considering its
       positional relationship or functional connection with the larger object.
    2. The larger object may contain another object with similar attributes to the
       main object. The caption should be written in a way that clearly distinguishes the
       main object from these similar objects.
    3. Keep the caption concise, limiting it to one sentence while ensuring 
       clarity and coherence.
    4. Do not explicitly mention the yellow contour or its presence in the image.
    5. Use contextual information from Image-1 to describe the main object's 
       relationship with the larger object, while referencing its attributes from Image-2.
    6. Contextual details from Image-1 and the description of the larger object
       should be used only to support the description of the main object.
</outputFormat>
<outputExamples>
    "8 in-context examples"
</outputExamples>
        \end{lstlisting}
    \end{minipage}
    \caption{Prompts for top-down generation. Captions are generated hierarchically from main objects to descendants while ensuring contextual coherence and uniqueness.}
    \label{tab:supp_prompt_top_down}
\end{table*}

%% file: tables/supplementary/prompt_bottom_up.tex
\definecolor{codebg}{rgb}{0.95, 0.95, 0.95}  
\lstdefinestyle{mystyle}{
    backgroundcolor=\color{codebg},   
    basicstyle=\ttfamily\small,       
    frame=single,                    
    breaklines=true,       
    breakautoindent=false,
    captionpos=b,                      
    keywordstyle=\bfseries,            
    commentstyle=\color{gray},         
    numbersep=5pt,                    
    xleftmargin=5pt, xrightmargin=5pt,
    prebreak={},  
    postbreak={}, 
}

\begin{table*}[t]
    \centering
    \captionsetup{type=table}
    \begin{minipage}{\textwidth}
        \begin{lstlisting}[style=mystyle]
<task>
    You are a detailed caption generator tasked with describing the main object in images.
    Your goal is to create precise and detailed captions while avoiding hallucination.
</task>
<objectives>
    1. The caption must primarily focus on the main object while considering its 
       contextual information to clearly identify what it is.
    2. The caption must emphasize the main object's attributes, such as color, texture,
       shape, and action if visible.
    3. Describe only what is visible in the image. Avoid adding any information that
       is not present.
    4. The main object is highlighted with a yellow contour.
    5. A short description of the main object will be provided in the prompt, which
       can be used to describe the main object.
    6. The main object consists of multiple subparts, and descriptions of these subparts
       will be provided in the prompt.
    7. The description of subparts may contain inaccurate, unimportant, or redundant
       information. Use only the essential details that do not contradict the 
       given image to ensure that the caption for the main object compositionally
       reflects relevant information from these subparts.
</objectives>
<inputDetails>
    1. An image with the main object marked by a yellow contour will be provided.
    2. A short description of the main object will be included in the prompt.
    3. Descriptions of the subparts of the main object will also be provided in
       the prompt.
</inputDetails>
<descriptionOfMainObject>
    "Description from the main object."
</descriptionOfMainObject>
<descriptionOfSubparts>
    "Descriptions from the child objects, if present."
</descriptionOfSubparts>
<outputFormat>
    1. Provide a single descriptive paragraph that focuses on the main object.
    2. Do not use bullet points or lists.
    3. Incorporate details from the provided descriptions to accurately depict the
       main object.
    4. Never mention the presence of the yellow contour in any form.
    5. Structure the caption clearly and concisely, avoiding excessive detail or
       verbosity. Do not start with phrases like "The image shows...".
    6. Ensure the focus is evident without explicitly stating that it is the main object.
</outputFormat>
        \end{lstlisting}
    \end{minipage}
    \caption{Prompts for bottom-up generation. Captions are refined by incorporating child node information to maintain hierarchical consistency.}
    \label{tab:supp_prompt_bottom_up}
\end{table*}

%% file: tables/supplementary/prompt_unique.tex
\definecolor{codebg}{rgb}{0.95, 0.95, 0.95}  
\lstdefinestyle{mystyle}{
    backgroundcolor=\color{codebg},   
    basicstyle=\ttfamily\small,       
    frame=single,                    
    breaklines=true,       
    breakautoindent=false,
    captionpos=b,                      
    keywordstyle=\bfseries,            
    commentstyle=\color{gray},         
    numbersep=5pt,                    
    xleftmargin=5pt, xrightmargin=5pt,
    prebreak={},  
    postbreak={}, 
}

\begin{table*}[t]
    \centering
    \begin{minipage}{\textwidth}
        \begin{lstlisting}[style=mystyle]
<task>
    You are a caption refinement model that enhances given descriptions to generate unique
    and precise captions for objects in an image. Your goal is to refine the provided
    caption based on contour-based indexing while maintaining clarity and specificity.
</task>
<objectives>
    1. Describe only what is visible in the image. Avoid adding any information that is
       not present.
    2. The image contains multiple contours in different colors, each with a
       corresponding index, marking distinct objects.
    3. The main object corresponds to index 0 and is specifically outlined with a
       blue contour.
    4. Your task is to refine the caption for index 0, highlighting its unique attributes
       while clearly differentiating it from other indexed contours in the image.
    5. The refined caption must primarily focus on index 0 while considering its
       contextual information to clearly identify it from other indices.
    6. The caption must emphasize index 0's attributes, such as color, texture, shape,
       and action, to make caption unique.
</objectives>
<inputDetails>
    1. The contours in the image are color-coded, and each contour has a
       corresponding index.
    2. The index corresponding to each contour is placed at the center of the contour,
       matching its color.
    3. The initial caption for index 0 (blue contour) is provided as input.
    4. The refined caption should ensure the distinction between index 0 (blue contour)
       and other objects in the image.
</inputDetails>
<refinementGuidelines>
    1. Preserve the core meaning of the given caption while improving its specificity
       and uniqueness.
    2. Emphasize key attributes that differentiate index 0 (blue contour) from
       other indices.
    3. Avoid mentioning the presence of contours or annotations explicitly in the caption.
    4. Keep the refined caption clearly yet descriptive.
    5. Ensure that the final caption remains a natural, human-like description of
       the object.
    6. Do not use bullet points or lists.
    7. Do not start the answer with words like "Certainly!".
</refinementGuidelines>
<captionForIndex0>
    "Description from the target (index 0) object"
</captionForIndex0>
<outputFormat>
    1. Provide a single descriptive paragraph that maintains clarity and coherence
       focusing on index 0 (blue contour)
    2. The refined caption should distinguish index 0 (blue contour) from other indices.
    3. Avoid generic or ambiguous descriptions.
    4. The refined caption should make index 0 clearly stand out from the other indexed
       objects without using phrases like "distinguished by" or similar expressions.
    4. Do not reference the contour colors or indices directly.
</outputFormat>
        \end{lstlisting}
    \end{minipage}
    \caption{Prompts for uniqueness refinement. Captions are refined by distinguishing visually similar regions while preserving semantic relevance.}
    \label{tab:supp_prompt_unique}
\end{table*}